\documentclass[10pt,twocolumn,letterpaper]{article}

\usepackage{cvpr}
\usepackage{times}
\usepackage{epsfig}
\usepackage{graphicx}
\usepackage{amsmath}
\usepackage{amssymb}
\usepackage{multirow}
\usepackage{algorithm}
\usepackage{algpseudocode}
\usepackage{amssymb,amsfonts}
\usepackage{amsmath,bm}
\usepackage{comment}
\usepackage{float}
\newcommand{\eat}[1]{}

\usepackage[breaklinks=true,bookmarks=false]{hyperref}

\cvprfinalcopy 

\def\cvprPaperID{****} 

\ifcvprfinal\fi
\begin{document}

\title{Defending Against Adversarial Attacks Using Random Forest}

\author{Yifan Ding$^1$ \hspace{0.4cm} Liqiang Wang$^1$ \hspace{0.4cm} Huan Zhang$^2$\hspace{0.4cm} Jinfeng Yi$^3$\hspace{0.4cm} Deliang Fan$^1$ \hspace{0.4cm} Boqing Gong$^4$ \\
$^1$University of Central Florida\\
{\tt\small yf.ding@knights.ucf.edu, lwang@cs.ucf.edu, dfan@ucf.edu}\\
$^2$University of California, Los Angeles \hspace{1cm}
$^3$JD AI Research\hspace{2cm}
$^4$Google \\
{\hspace{3cm}\tt\small huanzhang@ucla.edu \hspace{2cm}jinfengyi.ustc@gmail.com \hspace{0.1cm}boqinggo@outlook.com}
}


\maketitle

\begin{abstract}
   As deep neural networks (DNNs) have become increasingly important and popular, the robustness of DNNs is the key to the safety of both the Internet and physical world. Unfortunately, some recent studies show that  adversarial examples, which are hard to be distinguished from real examples, can easily fool DNNs and manipulate their predictions. Upon observing that adversarial examples are mostly generated by gradient-based methods, in this paper, we first propose to use a simple yet very effective non-differentiable hybrid model that combines DNNs and random forests, rather than hide gradients from attackers, to defend against the attacks. Our experiments show that our model can successfully and completely defend the white-box attacks, has a lower transferability, and is quite resistant to three representative types of black-box attacks; while at the same time, our model achieves similar classification accuracy as the original DNNs. Finally, we investigate and suggest a criterion to define where to grow random forests in DNNs.
\end{abstract}

\section{Introduction}

Despite being remarkably effective in solving many real-world problems such as the perception of images, videos, and texts, Deep neural networks (DNNs) are surprisingly vulnerable to adversarial attacks: one can easily fool the networks and manipulate their predictions by generating adversarial examples that are only slightly different from the real examples for which the networks usually give rise to correct predictions~\cite{papernot2016limitations}. While at the same time, DNNs are widely applied to many critical real-life applications, such as self-driving cars, robotics, and Internet of Things, it is vital to improve their robustness against adversarial attacks.

In the state-of-the-art adversarial attacking methods, gradients are the key ingredient to perturb a normal example to an adversarial one. These methods include white-box attacks~\cite{goodfellow2014explaining,goodfellow2014explaining}, which assume that attackers have full knowledge of the DNNs being attacked, including the architecture and weights of the DNNs and,  even the training data and gradients during training. In like manner, black-box attacking algorithms \cite{papernot2017practical,liu2016delving} allow attackers access to nothing but the DNNs' outputs (\eg  classification prediction probabilities) which are usually used to estimate the gradients. 
Thus recent defense strategies focus on preventing the attackers from inferring the gradients. 
As summarized in~\cite{athalye2018obfuscated}, defense methods often use shattered gradients~\cite{buckman2018thermometer}, design stochastic gradients, or intentionally vanishing/exploding gradients~\cite{song2017pixeldefend}. 
But since these methods are in the nutshell of training with back-propagation, it essentially opens the door to attackers. The gradient-based attack could still succeed as long as they find a way to obtain or approximate the correct gradients, which seems feasible in most cases~\cite{athalye2018obfuscated}.

In this paper, we open a novel direction for defending against gradient-based attacking models by introducing non-differentiable algorithms into DNNs. We propose a simple, intuitive yet effective method to transform a fragile DNN classifier into a more robust one. We present a hybrid approach that we transform the last few layers of DNNs classifiers into random forests that are non-differentiable at all so that the gradients cannot been approximated. Our proposed technique is very flexible for easily marrying the strength of the representation power of DNNs as feature extractor and the security property thanks to the non-differentiability of random forests. 

Extensive experiments on several DNN classificaton models including AlexNet~\cite{krizhevsky2012imagenet},  VGG-16~\cite{simonyan2014very}, and MNISTnet~\cite{carlini2017towards} show the effectiveness of our method against some popular attacking algorithms such as C\&W model~\cite{carlini2017towards} for white-box attacks and Zeroth Order Optimization (ZOO)~\cite{chen2017zoo} for black-box attacks. White-box attack methods fail to attack our hybrid deep model because the last few layers are not differentiable at all. Furthermore, most previous defense methods suffer from failing to defend the transferability of adversarial examples. We show that our approach using random forests leverages the problem. On the other hand, black-box attacks also find it hard to attack such hybrid model because the final prediction is not a single continuous-valued probabilistic output but an ensemble output. Details of experiments will be discussed latter.

We further provide technical study of our approach. We investigate the performance of building the random forests from different layers of the DNNs. We find that the relative $\ell_2$ distance between the real examples and the corresponded adversarial examples serves as good indicator to determine the depth of DNN layers that the random forests best grow upon. With careful examination and ablative study, experiments show our models not only defend against the attacks successfully but also achieve a high classification accuracy on the original test instances.



Below we summarize the key contributions:
\begin{itemize}
\item We present a simple, flexible, and intuitive technique to enhance the security of DNN classifiers against adversarial attacks.
\item We propose a novel direction for defending from gradient-based attacks. We utilize random forests to transform pure DNNs into non-differentiable classifiers thus prevent the gradient from being inferred which is the strategy that state-of-the-art attacking models typically rely on.
\item Extensive experiments show the effectiveness of our model against some strong attackers in both white-box and black-box literature.
\end{itemize}



\section{Related work}
\label{related}

Since the concept of adversarial samples comes up, many attack and defense methods have been proposed. 
In this section, we describe recent developments related to this topic, including both adversarial attacks and adversarial defense approaches.

\subsection{Adversarial attacks}
\textbf{White-box attacks}. White-box means attackers have the full access of the architecture, parameters and weights of the model~\cite{szegedy2013intriguing,carlini2017towards,goodfellow2014explaining}.  One of the first methods is proposed in~\cite{szegedy2013intriguing}, known as box-constrained L-BFGS, which minimizes the additive perturbation based on the classification constrains. 
 Then, Goodfellow {\it et al.} proposed an approach called Fast Gradient Sign Method (FGSM)~\cite{goodfellow2014explaining} to conquer the inefficiency. Later, Basic Iterative Method (BIM) method~\cite{kurakin2016adversarial} is introduced to add perturbations iteratively. In 2017, a very strong attack~\cite{carlini2017towards} ({\it a.k.a.} C\&W) method is proposed to find the minimal 
$L_0$, $L_2$, and $L_{\infty}$
distance. There are also generative network based attacking methods. For example,~\cite{baluja2017adversarial} generates adversarial samples through generative adversarial networks~\cite{goodfellow2014generative}. It is noteworthy that Kantchelian proposes an attack method on tree ensemble classifiers~\cite{kantchelian2016evasion}. However, the approach handles regression trees while we use classification trees, while the authors also propose that the attack of tree classifiers can be easily solved with adversarial training.


\textbf{Black-box attacks}\label{black}. Compared to white-box attack methods that request the target neural networks to be differentiable, black-box attacks are introduced to deal with non-differentiable systems or systems whose parameters and weights cannot be reached. Zeroth Order Optimization (ZOO)~\cite{chen2017zoo} can directly estimate the gradients of the target network through zeroth order stochastic coordinate descent. Papernot {\it et al.} propose ~\cite{papernot2017practical} to achieve decision boundaries learning based on transfer attack. And Liu {\it et al.}~\cite{liu2016delving} use an ensemble of several pre-trained models as the source model to generate adversarial examples even when no query of probes are allowed. 

 



\subsection{Adversarial defenses}

\textbf{Defenses based on gradient masking}. Gradient masking is one of the most popular defense methods that intentionally or unintentionally mask the gradient that is needed for computing perturbations by most white-box attackers. Buckman encodes input images using  thermometer encoding to enable discrete gradient that cannot be attacked directly~\cite{buckman2018thermometer}. While Guo applies transformation to the inputs to shatter the gradients~\cite{guo2017countering}. By adding noise to the logit outputs of neural networks, Nguyen introduces the masking based defense against C\&W attacks~\cite{nguyenlearning}. 


\textbf{Defenses through adversarial training}. Adversarial training is one of the most straightforward and efficient strategies when defending against the adversarial attacks~\cite{szegedy2013intriguing}. However, adversarial training only improves the robustness of some specific attacks~\cite{moosavi2016deepfool,biggio2013evasion}. For some strong attack methods like~\cite{carlini2017towards}, it is very hard to gain robustness via adversarial training because of large search space. It is observed that by adding noise before starting the optimizer for the attack~\cite{madry2017towards}, the over-fitting towards specific attacks can be reduced and ensembling adversarial samples generated by different models also improves robustness~\cite{tramer2017ensemble}.

\textbf{Other defenses}. There are also other defense methods that fall into none of the two categories above. Metzen defends against attacks through refusing classification by detecting whether there are signs of tampering of the input~\cite{metzen2017detecting}. Goodfellow uses shallow RBF that is robust to adversarial samples but  has a much lower accuracy on the clean samples~\cite{goodfellow2014explaining}. Some other methods try to apply preprocessing techniques towards the input to denoise before the classifier such as JPEG compression~\cite{das2017keeping} and median filter~\cite{xu2017feature}. 

\section{Our Approach}\label{approach}

To adversarially attack deep neural networks (DNNs), the most effective existing methods are almost gradients-based (\eg~\cite{carlini2017towards,goodfellow2014explaining}) probably because the DNNs are usually trained by (stochastic) gradient descent. As a result, to defend DNNs against such attacks, gradient masking has become the main idea behind recent defense methods~\cite{buckman2018thermometer}. It means that although the gradients exist in the deep models, they could be hidden from the attackers. However, such defenses fail once attackers find some ways to approximate the gradients~\cite{athalye2018obfuscated}. We first introduce our methdology, then summarize the the two streams of attack models we used to examine our method.
\begin{figure*}[t]
\begin{center}
\includegraphics[width=0.9\textwidth]{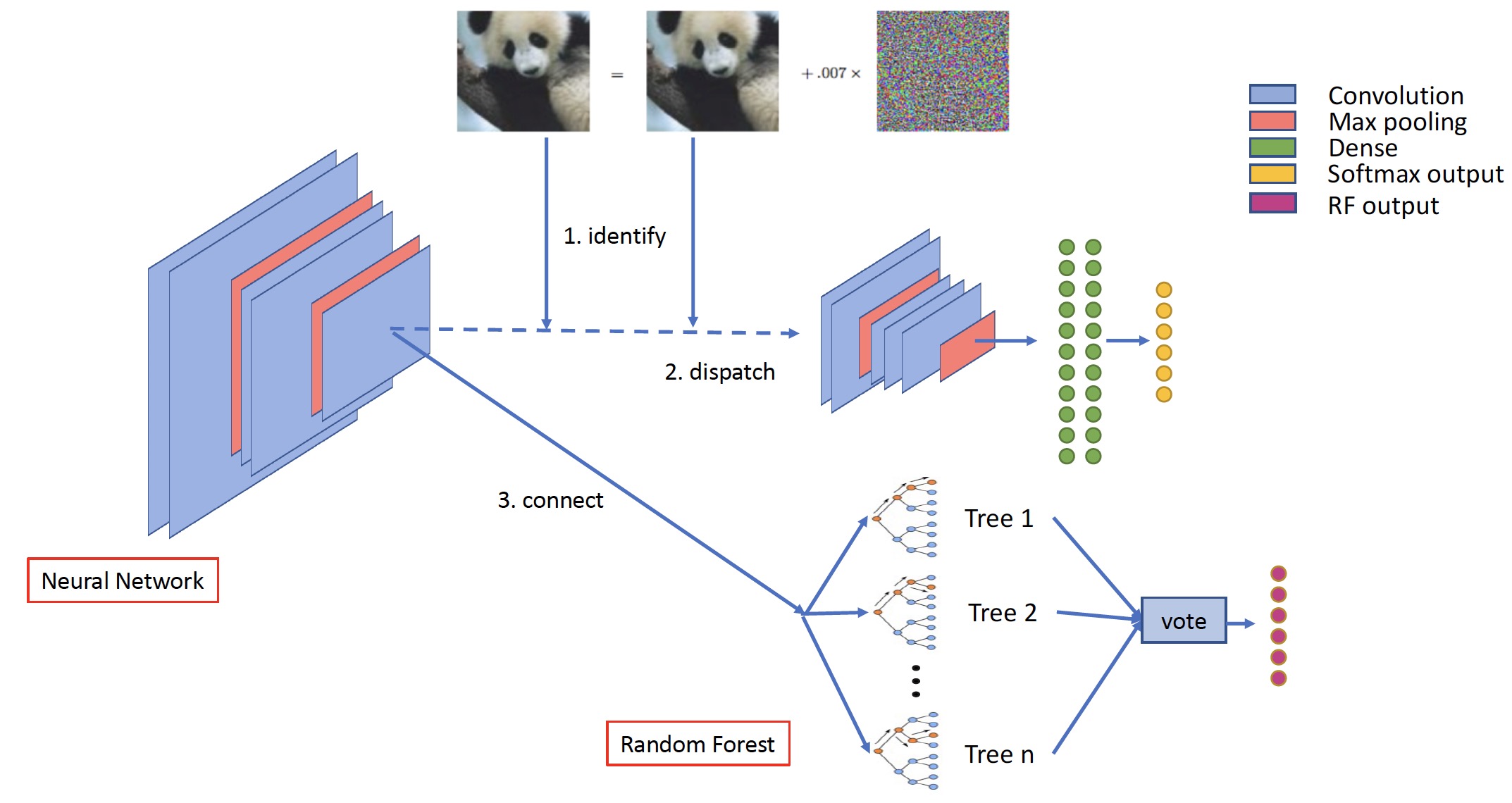}
\end{center}
   \caption{An overview of our approach. We build a hybrid model through three steps: First we identify the proper layer of growing the random forest by locating the $k$th layer in the DNN which has the biggest amplifying ability with Equation~\ref{eq:l2}. Then we dispatch the DNN from the $k$th layer and connect a random forest to imitate the classification ability of the DNN and dissolve the gradients to avoid being attacked by the adversarial attackers. 
   }
   \vspace{-10pt}
\label{fig:overview}
\end{figure*}
\subsection{Model}\label{definition}
We design a defense approach with a completely different strategy instead. By incorporating a standard CNN classifier with a random forest into a robust classifier, we integrate the merit of neural networks in extracting abstract and informative features and the advantage of non-gradient based classifiers. In practice, our hybrid model is achieved through a two-stage training procedure. We first train a DNN classifier, with back-propagation from cross-entropy loss, then discard the last few layers in the DNNs and replace them with a random forest ({\it e.g. built with classification trees}) to work as a non-differentiable classifier. Figure~\ref{fig:overview} illustrates our approach. It could be understood as attaching random forests upon a well-trained DNN feature extractor. Thanks to the powerful feature extractor and the expressive capabilities of random forests, our approach achieves almost the same accuracy as the original DNN classifier on the testing sets. Meanwhile, it disable white-box attacking methods to compute gradients through such a hybrid model. We show in experiments that our hybrid model is also robust to the transfer attacks which we proposes as the ultra white-box attack.

In formal, we introduce our hybrid model. To re-purpose pre-trained DNNs as feature extractors~\cite{donahue2014decaf}, a straightforward implementation is to use output from the penultimate layer. However, in practice directly stacking a random forest after the penultimate layer yields poor performance. To address this problem, we propose a passive strategy to identify the proper layer of the DNNs from which we grow the random forests. We find that after adding a small perturbation to the input, there is a difference for the relative $\ell_2$ distance between the original samples and the corresponding adversarial samples generated by the C\&W attack after each activation, which means that different layers in the DNNs have different amplifying ability towards the perturbation. The  average relative $\ell_2$ distance of the $k$th layer is calculated based on:
\begin{align}
D_k = \frac{1}{|M|}\sum_{i=1}^M{\frac{\|\bm{a}^{(k)}_i - \Tilde{\bm{a}}^{(k)}_i\|_2}{\|\bm{a}^{(k)}_i\|_2}}
\label{eq:l2}
\end{align}
where $M$ denotes the number of images,  $\bm{a}^{(k)}_i$ and $\Tilde{\bm{a}}^{(k)}_i$ denote the activation of the $i$th original input $x_i$ and the corresponding adversarial example $\Tilde{x}_i$ at the $k$th layer of DNN $\mathcal F$, respectively.

Using $\ell_2$ as the criterion, we keep as many layers as possible at the beginning from the input layer until we reach the one that causes significantly larger distortion than the other layers. Our experiments in Section \ref{experiments} demonstrate that this criterion is effective, allowing the resulting hybrid model to achieve a high classification accuracy and meanwhile to be robust against the transfer attacks or black-box attacks.

Afterwards, we use the output of the $(k-1)$th layer of the DNN $\mathcal F$, while $k$ denotes the layer with the largest relative $\ell_2$ distance between the original samples and the adversarial samples, as the input of the random forests. We build our random forests with classification trees by following the approach in~\cite{colthurst2016tensorforest}. For each split in each tree, we compute the split score $S$ through:
\begin{align}
S = G(L)+G(R)
\label{split_score}
\end{align}
where $L$ and $R$ denote the left and right per-class probability count vector, respectively. The idea is to calculate the Gini impurity to give the lowest split score $S$. The Gini impurity is calculated by:
\begin{align}
G(C) = 1 - \sum_{j}^J{(\frac{C_j}{N_c})^2},  
N_c = \sum_j^J{C_j}
\label{rf}
\end{align}
where there are J classes and $C_j$ denotes the number of samples classified into the $j$th class. The problem can be solved through Hoeffding tree~\cite{domingos2000mining}. Let $P_h$ denote the output of the $h$th classification tree and suppose there are $H$ trees in the random forests. The final predictions of sample $x$ in our hybrid model becomes the ensemble of the classification trees:
\begin{align}
P_f(j|x) = \frac{1}{H}\sum_{h=1}^{H}{P_h(j|\mathcal F_{k-1}(x))}
\label{rf}
\end{align}
where $\mathcal F_{k-1}$ denotes the output of the $(k-1)$th layer in the DNN $\mathcal F$, $j$ is the $j^{th}$ class and $H$ is the number of trees in the forest.

Next, we introduce the two categories of white-box and black-box attacks we mainly focus against in experiments.  









\subsection{Defending against white-box attacks and transfer attacks}\label{ultra}

White-box attacks are the most powerful adversarial attacks. However, by stacking the random forests, the hybrid model becomes non-differentiable and the conventional white-box attackers cannot get meaningful results by attacking the hybrid model because there is no way of back-propagating gradients through random forests.  Besides, to further demonstrate the robustness of the hybrid model, we also propose a attack setting named {\it ultra white-box attack}. In the ultra white-box attack setting, not only our our hybrid model but also the original DNNs upon which our model is built are released to attackers. Therefore, the ultra-white-box attackers will be able to generate adversarial examples for our end-to-end differentiable DNNs through gradient methods and transfer the attacking results to our hybrid model. The setting is based on the observations that adversarial examples have remarkable transferabilities~\cite{liu2016delving,szegedy2013intriguing}: given two neural networks trained for solving the same problem, the adversarial examples generated for one neural network can also successfully attack the other. 

In the white-box attacks experiments, a strong white-box attack, C\&W attack, is used to conduct the experiments. Specifically, we use their most efficient $\ell_2$ attack.
\subsection{Defending against black-box attacks}   

We also discuss the robustness of our model against the black-box attacks since black-box attackers should be able to attack any model without access to the gradients. As there are countless number of black-box attack methods and we would not be able to test all of them, we therefore categorize the existing black-box attacks into the following three representative groups. Then in each group, a representative and latest method is selected to test the effectiveness of our method. 

\paragraph{Gradient estimation}

One of the most efficient category is gradient estimation based methods that use various techniques  to estimate the unknown gradients of the target model~\cite{chen2017zoo,narodytska2017simple,bhagoji2017exploring}. We choose ZOO~\cite{chen2017zoo} out of this category in our experiments as it is one of the strongest black-box attack methods. ZOO achieves almost the same success rate as white-box attacks by estimating the gradients of the target network through the zeroth-order stochastic coordinate descent.
It is worth pointing out that there is no ground-truth gradients at all in our hybrid  model. However, ZOO  still tries to attack it by the zeroth-order gradients. As a result, our approach is experimentally shown hard to be attacked by ZOO.
\begin{table*}[h]
\caption{The average relative $\ell_2$ distance between adversarial and original samples}\label{tab:l2}
\begin{center}
\begin{tabular}{lccccccccccccccc}
\hline
 Activation  &1 & 2  & 3&4&5&6&7&8&9&10&11&12&13&14&15  \\
\hline\hline
MNISTnet & 0.03 &0.05&0.08 & 0.18& 0.22&  0.30& \textbf{0.60}&-&-&-&-&-&-&-&-\\
AlexNet &0.01 &  0.03 &  0.08 & 0.17 &0.17  & \textbf{0.43}& 0.50&0.62 &-&-&-&-&-&-&-\\
VGG-16 & 0.01 &  0.02 &0.03& 0.07 & 0.12& 0.14  &\textbf{0.29} & 0.43 &0.45  &0.51 &0.53& 0.48 & 0.53 &0.52 &0.65\\
\hline
\end{tabular}
\end{center}
\end{table*}
\vspace{-5pt}
\paragraph{Decision-based attacks}\label{ensemple}

The decision-based attacks are black-box methods that solely rely on the final decision of the model~\cite{nelson2012query,biggio2013evasion}. As a representative decision-based method, Boundary Attack~\cite{brendel2017decision} achieves a comparable attack performance with one of the best gradient-based attacks~\cite{carlini2017towards} in both targeted and untargeted scenarios which literately modifies a random sample from the target class to be visually similar to the original sample.
Boundary Attack is hard to be defended because one of the pre-requisites of the algorithm is to maintain the target classification. Therefore, the attack success rate would always be 100\% although the generated adversarial samples might be quite different from the original samples. During the experiment, we use the $\ell_2$ distance to evaluate the performance of the attack and defense.

\paragraph{Training a substitute model}
While knowing the number of target classes, the attackers could train a substitute DNN to mimic the behavior of the model. Ideally, the substitute DNNs shares the same decision boundaries as the target model and the adversarial samples generated by the substitute DNNs would be able to fail the target network as well. We use the recent Practical Black-Box Attack method proposed in~\cite{papernot2017practical} to test our defending approach against this type of methods. 

\section{Experiments}\label{experiments}

Our experiments are conducted on CIFAR-10~\cite{krizhevsky2009learning} and MNIST~\cite{lecun1998gradient} dataset, following the previous experiment protocols~\cite{carlini2017towards,chen2017zoo}.  We pre-process all the images by re-scaling the pixel values to the range [-0.5, 0.5]. For all of the experiments on CIFAR-10, we use two popular neural networks VGG-16~\cite{simonyan2014very} and AlexNet~\cite{krizhevsky2012imagenet} as our backbone networks. For MNIST, we use the architecture proposed in the C\&W attack \cite{carlini2017towards}, {\it i.e.}, a 7-layer convolutional neural network and we denote it as {\it MNISTnet} in our following experiments. In the testing stage, we also follow the C\&W attack and use the first 1000 originally correctly classified images from the test set to evaluate the success rate of the attack. 


We consider the untargeted attack for the substitute model method~\cite{papernot2017practical} and targeted attack for the other methods. In the targeted attack, we use all the 9 classes in turn except for the ground truth class as the target class for each image and calculate the average attack success rate over all the target classes. For the ZOO and ultra white-box attacks, margin $\kappa$~\cite{carlini2017towards} is set to zero. 
In our experiments, we use the attack success rate (\ie ASR) to measure the effectiveness of different attacks (or, our defense).  ASR is calculated as  the percentage of the adversarial examples that are successfully classified to  the target classes by the attacks out of all examples tested.  Additionally, we also include $\ell_2$ distance in most of our experiments~\cite{carlini2017towards,chen2017zoo}, which is calculated using the mean squared distance  between the original images and the adversarial images. We find that a random forest of 300 trees and 10,000 nodes gives rise to the best trade-off between the classification accuracy and the defense performance on CIFAR-10.  However, to save computational cost, we use 300 trees and 10,000 nodes for AlexNet and 300 trees and 1,000 nodes for VGG-16 in our experiments. For AlexNet, we keep the first five layers as the feature extractor and, for VGG-16, we grow the forest from the $8^{th}$ layer.  For the neural network applied to MNIST, we use 10 trees and 1,000 nodes taking as input the activations of the $6^{th}$ layer.





\subsection{Finding the proper layers to grow the random forests}\label{sec:sub}
To find the most suitable layer to grow the random forests, we use the relative $\ell_2$ distance (Equation~\ref{eq:l2}) defined in Section~\ref{definition} as our criterion.  In this experiment, we randomly select 1,000 originally correctly classified images out of the validation sets to evaluate the relative distortion between the original samples $\{x\}$ and the adversarial samples $\{\Tilde{x}\}$ generated by the C\&W attack.
Since the amplifying ability of each activation is different, the relative growth of $\ell_2$ distance between each two activations changes. To our surprise, the relative $\ell_2$ distance grows slowly in both the first layers and the last few layers for AlexNet and VGG-16, while on the contrary, faster in some intermediate layers as shown in Table~\ref{tab:l2}.

Recall that our original intention of building the hybrid model is to take advantage of the strength of the representation power of DNNs as feature extractor, the experimental results declare the fact that each layer in the neural networks plays a different role (\eg the bottom layers extract low level features and the top layers in charge of the classification). Then the different role of each layer result in the different ability of the layer to manipulate the adversarial distance. It turns out the intermediate layers in AlexNet and VGG-16 have best of the ability to generate the adversarials. Hence we are safe to replace the top layers which are responsible to classification with the random forest which has no gradient. The distance also grows differently in different networks (MNISTnet and others), which also request special dispatch point in distinctive networks.

We therefore choose to grow the random forests after the one with the largest relative growth which is the $6$th for MNISTnet, $5$th for AlexNet and $8$th for VGG-16 after trade-off between the defense performance and the classification accuracy. The selection turns out to be effective in the following experiments. The ablation study of the ultra white-box performance while growing random forests from different layers is shown in Section~\ref{substitute}.




\subsection{White-box attacks and transfer attacks}

We use the C\&W method as the white-box attack methods following the setting of the original paper~\cite{carlini2017towards}. As mentioned before, directly attacking our hybrid model gives 0\% attack success rate. And our model also has a very low transferability in the ultra white-box setting. Since there's no counterpart in the original DNNs for comparison with our ultra white-box setting, we also test the regular transfer attacks in which we generate adversarial samples on one network(\ie AlexNet) and test the accuracy of the generated adversarial samples on another network(\ie VGG-16 and our AlexNet hybrid model) The results are shown in Table~\ref{tab:ultra}. For this and the following tables, we denote accuracy as 'acc' and attack success rate as 'ASR'.


\begin{table}[h]
\caption{Results of the white-box and transfer attack}\label{tab:ultra}
\begin{center}
\begin{tabular}{l c c c c }
\hline

  
& & Acc\%  &  ASR\% &  Ultra\%\\
\hline\hline
\multirow{2}{*}{AlexNet} &dnn&83.4 &100&100\\
&ours&82.7 &\textbf{0.0} &\textbf{4.8}\\
\hline
\multirow{2}{*}{VGG-16}&dnn &93.5 &100&100\\
&ours&90.6 &\textbf{0.0}&\textbf{9.7}\\
\hline
\multirow{2}{*}{MNISTnet }&dnn&99.2 &100&100\\
&ours&98.4  &\textbf{0.0}& \textbf{4.4 }\\

\hline
\end{tabular}
\end{center}
\end{table}

Compared to the 100\% attack success rate achieved in the original DNNs, the C\&W attack barely succeeds in one single attack to our hybrid models as shown in the $4$th column of Table~\ref{tab:ultra}. Besides, testing the transferability on the same DNNs is meaningless, we just keep it as 100\% for comparison. After making it a hybrid model, we achieves a transfer attack success rate of 4.8\%,9.7\% and 4.4\% on AlexNet, VGG-16 and MNIST, respectively. Thus we can safely claim that even if we release the architecture and weights of both our original neural network and the hybrid network, our approach can still successfully defend against the current strongest white-box attack.



\subsection{Black-box attack: ZOO}

Since the comparison is already significant enough, in the experiment of the Zeroth Order Optimization (ZOO) attack~\cite{chen2017zoo} we only run 1,500 iterations to attack the original neural networks but 15,000 iterations to attack our hybrid model to save computational time. For MNIST, we did not re-produce the experiment but directly include the result reported in their paper as the same network is used. The results are shown in Table~\ref{tab:zoo}.

\begin{table}[h]
\caption{Results of defending against the ZOO attack~\cite{chen2017zoo}}\label{tab:zoo}
\begin{center}
\begin{tabular}{ l c c c c }
\hline
Network &&Acc\%& ASR\% & $\ell_2$ distance\\
\hline\hline
 
\multirow{2}{*}{AlexNet} &dnn&   83.2 &89.7 &8.79  \\
&ours&82.7 &\textbf{5.4 }&\textbf{37.2}\\
\hline
\multirow{2}{*}{VGG-16}&dnn  &93.5 &86.8  &9.89  \\
&ours&90.6 &\textbf{3.6} &\textbf{38.1}\\
\hline
\multirow{2}{*}{MNISTnet }&dnn&99.2 &98.9  &2.0 \\
&ours &98.2 & \textbf{0.9}&\textbf{185.2}  \\
\hline
\end{tabular}
\end{center}
\end{table}


In terms of the classification accuracy, our hybrid models are on a par with the original neural networks; the absolute decreases are only 0.5\% 2.9\% and 1\% for AlexNet VGG-16 and MNISTnet, respectively. In terms of the defense performance, the hybrid models are much better than the original neural networks.  For AlexNet, the attack success rate (\ie ASR, the lower the better) decreases from 89.7\% to 5.4\% with an average $\ell_2$ distance between the original input examples and the adversarial examples increasing from 8.79 to 37.2 (the higher the better from the defense perspective). The same observation also happens to VGG-16 and MNISTnet, where the ASR decreases from 86.8\% to 3.6\% and from 98.9\% to 0.9\%  with an average distance increasing from 9.89 to 38.1 and from 2.0 to 185.2, respectively. Thanks to  our hybrid defense method that has about the same classification accuracy as DNNs, we can safely claim that the strong black-box attack method, ZOO,  fails.

\subsection{Black-box attack: Decision-based attack}
For decision-based attack~\cite{brendel2017decision}, we test 100 samples from the testset that have been correctly classified by our model. For each example with label $l$, we set the target label as $(l+1) \mod 10$. For each example, we set the maximum number of steps to be 100, where very few steps are already sufficient to demonstrate the difference between our model and the original.  In each step, the length of the total perturbation $\delta$ and the length of the step $\epsilon$ towards the original input are initially set to 0.1 and 1.0, respectively, as suggested by \cite{brendel2017decision}. In each $\delta$ step, 5 orthogonal perturbations are generated compared with 10 in the original paper. Our results are shown in Table~\ref{tab:decision}.
\begin{table}[h]
\caption{Results attacked by Decision Based Attack Model~\cite{brendel2017decision}}\label{tab:decision}
\begin{center}
\begin{tabular}{ l c c c c}
\hline
Network &&Acc\% & $\ell_2$ dist & Time~(s)\\
\hline\hline
 
\multirow{2}{*}{AlexNet}&dnn &   83.2   & 4.58  & 1.84\\
&ours&82.7&\textbf{5.87}& \textbf{61.2}\\
\hline
\multirow{2}{*}{VGG-16}&dnn &93.5  & 5.66 & 2.57\\
&ours&90.6 &\textbf{5.74}&\textbf{46.0}\\
\hline
\multirow{2}{*}{MNISTnet }&dnn&99.2   & 16.6& 0.86\\
&ours &99.2 &11.1&\textbf{5499.6}\\

\hline
\end{tabular}
\end{center}
\end{table}

Since the decision-based method follows the rule that the target classification must been maintained, it makes no sense to compare the attack success rate. Instead, we compare the $\ell_2$ distance of the generated adversarial samples through attacking the DNNs and our hybrid models and the attack time. The longer the attack time and the $\ell_2$ distance, the poor the attack. Our results are shown on Table~\ref{tab:decision}.
 


\subsection{Black-box attack: Practical Substitute Model}

Recall that the substitute model in \cite{papernot2017practical} only conducts experiments on MNIST. We expand their experiment setting to CIFAR-10 by re-adjusting the parameters. To compare adversarial attacks on CIFAR-10, we set learn rate = 0.001. In addition, we increase the number of seed samples from 150 to 1500. 
The substitute model is also enhanced, specifically, we use VGG-16 as the substitute model for AlexNet and AlexNet as the substitute model for VGG-16. We use learning rates 0.001 and 0.0001 for AlexNet and VGG-16, respectively. We train the model with 5 data augmentations, each for 50 epochs, which is 1 data augmentation less than the original experiments in \cite{papernot2017practical} since the number of training samples grow exponentially with each data augmentation process and the seed training set is already 10 times larger compared to the original experiments. And for the experiment of MNIST, we use the same setting as the original paper, which means $\epsilon$ is set to 0.4. The results are shown in Table~\ref{tab:sub}. "Acc" and "Acc:sub" denote the accuracy of the original and the substitute model respectively.

\begin{table}[h]
\caption{Results attacked by Practical Substitute Model~\cite{papernot2017practical}}\label{tab:sub}
\begin{center}
\begin{tabular}{ l c c c c c c}
\hline
Network& $\epsilon$&  &Acc\% &Acc:sub\% &ASR\% \\
\hline\hline

\multirow{4}{*}{AlexNet}&\multirow{2}{*}{0.03}&dnn &   83.2&45.6 &\textbf{19.2} \\
&   &ours&82.5 & 44.1&19.8 \\
&\multirow{2}{*}{0.1}&dnn&83.2 &45.5&47.2  \\
&&ours&82.5 &43.0&\textbf{44.9}  \\

\hline
\multirow{4}{*}{VGG-16}&\multirow{2}{*}{0.03}&dnn &   93.5 &42.8&  13.1 \\
&&ours&90.5 &44.4 & \textbf{12.5} \\
&\multirow{2}{*}{0.1} &dnn&   93.5  &42.1 &70.3   \\ 
&&ours& 90.5  & 40.1&\textbf{69.7}   \\

\hline
\multirow{2}{*}{M-net}&\multirow{2}{*}{0.4}&dnn &   99.2 &70.1& 45.7 \\
&&ours &   98.7 &78.1&\textbf{43.1} \\
\hline
\end{tabular}
\end{center}
\end{table}

In Table~\ref{tab:sub}, ASR denotes the performance of our defense of untargetd attacks, as only untargeted attacks are involved in \cite{papernot2017practical}. It represents that the generated adversarial samples cannot be correctly classified by our hybrid model after the attack. It can be found from the experiment that it is very hard to train a substitute model on natural images especially when the model is complex. Therefore we also test when $\epsilon = 0.1 $ where $\epsilon$ is the perturbation magnitude preset for the FGSM attack~\cite{goodfellow2014explaining} which in most cases, $\epsilon$ is set to 0.031.


Among the three categories of black-box attacks, our hybrid model also performs well qualified. It either decreases attack success rate or increase $\ell_2$ distance, or attack time. To the best of our knowledge, there are very few research that have done a wholesome attack towards a single defense method from white-box to black-box attacks, from directly attacks to transfer attacks, and gain a satisfying performance on almost all of the experiments.

\section{Technical study}\label{ablation}

In this section, we perform technical studies to quantify the details of our designs including the effect of parameters involved in our model, the influence of adversarial training, and the difference of the depths of DNN layers that the random forests build on. Specifically, we re-train our approach using different parameters in the random forests and DNNs, and demonstrate the results after adversarial training. 

\subsection{Effectiveness of adversarial training}

To investigate whether adversarial training helps, we implement two kinds of experiments to verify this issue, as shown in Table~\ref{tab:adv}. In the first experiment, we train the original DNNs with the adversarial samples generated from C\&W attack and PGD attack~\cite{madry2017towards}, respectively, and then generate two new hybrid models. Then we test the two new hybrid models using the ultra white-box attack. It can be found from Table~\ref{tab:adv} that adversarial training on original DNNs does not affect the attack success rate too much but the average distortion increases.

\begin{table}[h]
\caption{Results of adversarial training}\label{tab:adv}
\begin{center}
\begin{tabular}{l c c   c  c}
\hline
   Adv train& Acc\% & ASR\% & Distortion\\
\hline\hline
None&83.4 &   32.8  &  0.44  \\
\hline
Hybrid:~C\&W & 81.1&33.5&\textbf{0.69}\\
Hybrid:~PGD & 79.4&33.9&\textbf{0.52}\\

Random~forest &83.2   &\textbf{16.2} &0.44\\

\hline
\end{tabular}
\end{center}
\end{table}
In the second experiment, for a hybrid model, we tune its random forest by the adversarial samples generated from C\&W attack. 
Then we test the tuned hybrid model again using the ultra white-box attack. The attack success rate drops since the random forests already learns the adversarial samples. To better observe the experiment difference, only the last few fully connected layers are deprived and replaced by random forests. The experiment is conducted on AlexNet and all the attack success rates are calculated based on the ultra white-box settings.


\begin{table*}[t]
\caption{Results of the ultra white-box attack with different number of trees and nodes}\label{tab:forest}
\begin{center}
\begin{tabular}{l l| c c c  c| c c c  c|c c c c}
\hline

  Trees  &  Nodes & \multicolumn{4}{c}{AlexNet}&\multicolumn{4}{c}{VGG-16}&\multicolumn{4}{c}{MNISTnet} \\
  
  & & Acc  &  ASR & Hyb acc  &Trans&Acc  &  ASR  & Hyb acc   & Trans&Acc  &  ASR  & Hyb acc   & Trans\\
\hline\hline
10&100&\multirow{9}{*}{83.4} &\multirow{9}{*}{100}&74.2  &10.7&\multirow{9}{*}{93.5} &\multirow{9}{*}{100}&83.1  &14.2&\multirow{9}{*}{99.2} &\multirow{9}{*}{100}& 94.0  &9.1 \\
10& 1000& & &77.7 &8.2 &  &  &86.4 &13.0&& &97.1&8.3\\
10& 10000& & &77.8 &8.9 &  &  & 86.8 &12.3&&&97.8&7.6\\
100&100&& &80.5 &6.2 &  &  & 87.9 & 11.8&&&96.4&7.1\\
100&1000& & &81.9 &6.2&  &  &89.8  & 11.6&&&97.8& 5.6 \\
100&10000& & &82.6 &5.8 &  &  & 90.4 & 10.4&&&98.2&4.5\\
300&100& &&80.8 &6.6 &  &  & 88.3 &10.7 &&&96.3&6.5\\
300&1000& & &82.3 &5.3  &  &  & 90.1 &11.1&&&97.7&5.1\\
300&10000& & &82.7 &\textbf{4.8 }& &  &90.6  &\textbf{9.7 }&&&98.4&\textbf{4.4}\\

\hline
\end{tabular}
\end{center}
\end{table*}
\begin{table*}[h]
\caption{Results when starting the substitution from the second-last and the best activation }\label{tab:ab_transfer}
\begin{center}
\begin{tabular}{l l| c c| c c |c c| c c}
\hline

  Trees  &  Nodes & \multicolumn{4}{c}{AlexNet}&\multicolumn{4}{c}{VGG-16} \\
  
      &   & \multicolumn{2}{c}{From $7th$ activation}&\multicolumn{2}{c}{From $6th$ activation} & \multicolumn{2}{c}{From $14th$ activation}&\multicolumn{2}{c}{From $7th$ activation}\\
  
  & & Hybrid acc  &  Trans & Hybrid acc  &  Trans& Hybrid acc  &  Trans & Hybrid acc  &  Trans\\
\hline\hline
10&100&81.4&\textbf{36.1}&74.2 &10.7&93.4& \textbf{48.6}&69.2 &9.0\\
100&1000&83.3&37.1&81.9&6.2&93.5 &55.2 &81.8 & 6.6 \\
300&10000&83.3&36.9&82.7&\textbf{4.8 }& 93.6&57.3 &84.4 & \textbf{5.6}\\
\hline
\end{tabular}
\end{center}
\end{table*}
\begin{table*}[h!]
\caption{Results when starting the substitution from different layers (AlexNet)}\label{tab:alexpool}
\begin{center}
\begin{tabular}{l l |c c| c c| c c| c c| c c}
\hline

  Trees  &  Nodes & \multicolumn{2}{c}{After the conv}&\multicolumn{2}{c}{After activation}&\multicolumn{2}{c}{After Pooling}&\multicolumn{2}{c}{After Batchnorm} \\
  
  
  & & Hybrid acc  &  Trans & Hybrid acc  &  Trans& Hybrid acc  &  Trans& Hybrid acc  &  Trans\\
\hline\hline
10&100&72.8 &13.0 &74.2  & 10.7& 75.9 & 17.7& 76.0&16.6  \\
100&1000&81.8&\textbf{9.8} & 81.9 & 6.2& 82.1 &\textbf{10.7} &82.2 & 11.2   \\
300&10000&82.6&10.6 & 82.7 &\textbf{4.8 }&83.0  &10.9 &82.9 &  \textbf{10.9}   \\

\hline
\end{tabular}
\end{center}
\end{table*}  
\subsection{Parameters in the random forests}\label{sec:random}

We only tune two parameters of the random forests, {\it i.e.,} the number of trees and the number of max nodes in each tree. We test the number of trees between 10 to 300 and the number of max nodes between 100 and 10000. The parameters are tested using grid search and the results are reported with respect to the performance of the ultra white-box task. Table~\ref{tab:forest} shows the results on AlexNet where the substitution is conducted from the $5^{th}$ layer and VGG-16 where the substitutition is conducted from the $8^{th}$ layer.  The activation layers are numbered from the bottom layers to top layers.

It can be found that the more trees and nodes we have, the higher the performance of our hybrid network achieves on the original testset, and the lower transferrability the hybrid network has. Basically, the defense performance grows with the performance of the hybrid model. 
This turns out to be a very pleasant trend since some existing defense models increase the robustness of their systems but at the cost of the drop of the accuracy~\cite{goodfellow2014explaining}. 
However, when the number of parameters grows, the system turns out to be slower and requires more computing resources. We utterly choose 300 trees and 10000 nodes to attain the proper performance and speed within our hardware capabilities.

\subsection{From which layer to grow random forests}\label{substitute}

Theoretically speaking, we can substitute the neural network from any layer. But actually, the starting layer matters regarding the final performance of our model. Using the proposed criterion defined in Section~\ref{sec:sub}, we find starting from the layer that has the largest gap of relative $\ell_2$ distance from its prior layer gains a better transferability than the penultimate layer, which is shown in Table~\ref{tab:ab_transfer}.


We also investigate the impact of max pooling and batch normalization layers. We conduct the experiment after the $6^{th}$ activation of AlexNet, where there are 4 layers including convolutional, activation, max pooling and batch normalization. Table~\ref{tab:alexpool} shows the best performance is achieved after the activation layer. 


\section{Conclusion}\label{conclusion}

To the best of our knowledge, there is no previous work considering using models with no gradients to defend adversarial attacks, probably because of the poor performance of non deep learning methods on high-level tasks (\eg image classification). While in our paper, we propose a simple hybrid model to overcome the drawbacks of individual DNNs and random forests. Through comprehensive experiments, we demonstrate that our proposed method is very effective for defending both white-box (including ultra white-box attack) and different kinds of black-box attacks. Simple but advantageous, our approach provides a preliminary evidence in support of the effectiveness of hybrid models, shedding lights on a promising direction to pursue.

{\small
\bibliographystyle{ieee}
\bibliography{egbib}
}
\end{document}